# Principles of modal and vector theory of formal intelligence systems

**Yuri Parzhin**
(*National Technical University "Kharkiv Polytechnic Institute" Ukraine*)

***Abstract***

*The paper considers the class of information systems capable of solving heuristic problems on basis of formal theory that was termed modal and vector theory of formal intelligent systems (FIS). The paper justifies the construction of FIS resolution algorithm, defines the main features of these systems and proves theorems that underlie the theory. The principle of representation diversity of FIS construction is formulated. The paper deals with the main principles of constructing and functioning formal intelligent system (FIS) on basis of FIS modal and vector theory. The following phenomena are considered: modular architecture of FIS presentation sub-system, algorithms of data processing at every step of the stage of creating presentations. Besides the paper suggests the structure of neural elements, i.e. zone detectors and processors that are the basis for FIS construction.*

## 1. Relevance

The stated problem is quite pressing nowadays because at present time there does not exist any unified theory that deals with the definition of algorithmic or formal systems capable to solve smart problems a human being can solve.

The examples of such problems are the following:

- external information perception and its interpretation in the terms of internal language of an intelligent system;
- classification and identification (including associative one) of accepted information, for example, objects, scenes, processes;
- definition of cognitive space-time environmental model based on the results of recognition;
- setting goals (generating hypotheses) and their solution (proof of hypotheses) using cognitive space-time model;
- implementation of communicative functions of interaction with other systems;
- system teaching with the use of other systems and self-teaching.

Modern math-based methods used for models of knowledge and data presentation (e.g. semantic networks, artificial neural networks, etc.) make it possible to solve only certain intelligent problems concerning definite knowledge domain that cause modern crisis of investigations in the field of artificial intelligence development.

Formal systems solving intelligent problems must meet the main requirement of denying any special math-based methods use. Meeting this requirement is necessary to develop an adequate model of human brain data processing as brain neural structures do not do any mathematic calculations. This does not contradict the statement that any digital system uses simple arithmetic operations to realize microprogramming control or that digital modeling of biophysical or biochemical processes which underlie in the basis of data processing by individual neurons also demand using elementary math-based procedures.

The objective of this research is to develop the main propositions of the theory named modal and vector theory of formal intelligent systems.

## 2. The main results of scientific research

*Formal intelligent system* (FIS) will be considered as data system capable to solve independently intelligent problems on the basis of a given set of formal rules as a result of automatic objective generation of well-formed formulae (WFF), i.e. hypotheses that have semantic interpretation, their proving or refuting as a result of the use of effective resolving algorithm without specific math-based methods.

*Effective resolution algorithm* (effective algorithm) will be considered as an algorithms capable over finite number of steps for any WFF to define if it is a theorem or not (if it is a part of the set of true formulae $T$ or a part of the set of false formulae $R$), i.e. capable to prove or disprove a hypothesis.

The suggested theory is based on the hypothesis by Vernon B. Mauntcastle [1] stating that brain or neocortex has modular hierarchical arrangement, all modules having similar structure and performing the same translation algorithm. This hypothesis is proved by a lot of modern neurobiological researches.

Each module of any perceptual or motor system processes data obtained, as a rule, from a set of data system modules that are placed on the lower hierarchical level (this data stream will be called *the main* or *upstream*), as well as from definite group of modules that are placed on the higher hierarchical level (this data stream will be called *downstream* or *control*). Besides, there exists horizontal interaction of modules that lie on the same level of hierarchy within definite areas (e.g. receptive fields). An evident result of each module operation is the convolution of data upstream up to a set of definite reactions of individual neurons – detectors at the modular output.

FIS are made from general-purpose modules functioning according to the single data processing algorithm. Module-making is based on the modern neurobiology understanding of integrative mechanisms that occur in neocortex, and, in particular, in visual cortex [2]. The approach will be described without exact reference to cytoarchitecture of neocortex areas that are studied.

### 2.1 Mathematical grounds of the theory

FIS as any other open system reflects actual external objects, processes, phenomena and their internal conception in FIS language alphabet symbols.

*Definition 1.* FIS subsystem that perceives data about actual objects, processes, and phenomena in parallel-serial input data stream and reforms it in FIS language alphabet symbols to develop data upstream will be called *subsystem* or *perceptual system* (PS).

Later in FIS continuous data upstream is segmented, i.e. every perceived object, process, and phenomenon is matched with definite expression – sequence of FIS language symbols. This sequence is initial internal conception of corresponding objects, processes or phenomena. This initial conception will be called *presentation*.

*Definition 2.* FIS subsystem that deals with presentations will be called presentation subsystem or system (PSS).

Presentations are linked in data upstream and make the *system of presentations*.

The adequacy of input data stream mapping into the system of presentations is relative because its completeness depends on PS characteristics (technical features of perception detectors, PS architecture, etc.), as well as on the architecture and algorithm PSS functioning. It is possible to state that any PS and PSS reflects input data stream into the system of presentations in part. But the result of each individual image – each presentation must be essential and sufficient to gain the aims of FIS functioning, and the process (algorithm) of presenting must be objective, i.e. independent from technical characteristics or features FIS architecture.

The most general objective of FIS functioning that makes it different from other formal systems is problem statement in the FIS language linked with automatic generation of hypotheses and quality decision-making in the terms of the given language under conditions of uncertainty or incomplete information of so called 'heuristic solutions' on the basis of efficient resolution algorithm. In fact the term 'heuristic solutions' only reflects the degree of our lack of knowledge of general formal approach to solving similar problems.

The key notion of any formal theory is the idea of '*truth*'.

There exist a lot of philosophic concepts, approaches, and definitions of 'truth'. The problem of interpretation of this notion is one of the most popular topics of not only antique but modern philosophy. In respect to FIS we are going to rely on the following interpretation of classic definition of 'truth'.

*Definition 3. Truth in FIS* will be considered as the mapping of perceivable objects, processes, and phenomena of environment in the system of presentations based on objective formal procedure (efficient algorithm).

Thus, if such formal procedure is probable, it is possible to define the following fundamental for this theory axiom.

*Axiom 1.* Any presentation in FIS PSS is true.

So there are no false presentations in PSS. Really, any open information system perceives environment and creates its relatively true model (presentation) more or less adequate.

Thus, the term 'false' that opposes the term 'true' cannot be used in FIS presentations. Presentations are facts, statements, and axioms of FIS PSS.

What is the sense of the notion 'false'? What does it exist for and how is it formed? To answer these questions it is necessary to pass on to more formal reasoning.

To simplify the description some limitations in the studied FIS must be defined. These limitations mainly deal with the types of environment perceived by FIS.

Let us suppose that in the environmental model there exist only contour 2D objects without internal structure. Any contour is presented by continuous sequence of dots, each being perceived as one bit of information with the meaning of "1". Objects are placed on the surface or white background, each dot of which being perceived as "0". Let us suppose that there exist a great number of such surfaces which simulates spatial structure called '*contour environment'* (CE). CE objects can participate in different processes. Let us suppose that FIS PS is presented by optic system with the matrix of electron detectors modeling receptive field of visual perception. In this case PS is defined as formal system.

Let us consider FIS PSS as a formal system $z_1$.

Any formal system is built on the basis of formal (axiomatic) theory [3].

In a general case, formal system $Z$ over alphabet $K$ is called the set of the following varieties:

1. $K$ alphabet;
2. $V$ set of language formulae construction rules WFF;
3. $A$ set of true language formulae is axioms;
4. $C$ set of true WFF deduction rules:

$$Z=<K,V,A,C> \qquad (1)$$

$K_1$ alphabet of $z_1$ system is a set of PS detector reactions. If PS detector matrix has n x n elements dimension, then $K_1$ has $n^2$ symbols of "1", their feature being their position in the matrix. Let us set implication sign ($\rightarrow$) as well.

Let us define $V_1$ set of WFF.

1. $a_i = 1$ , where $i$ – is the number of symbol in $K_1$ alphabet – WFF. Any sequence of alphabet symbols is called an expression.
2. Let $F_i(x_i)=(a_1,a_2,\dots,a_m)$ be a $x_i \in X_1$ definite characteristic linear ordered set of symbols, where $a_1$ is initial, $a_m$ – final symbols in the given sequence, i.e. $a_0$ and $a_{m+1}$ symbols have "0" meaning. Then $F_i(x_i)$ is WFF.

*A set of* $X_1=(x_1,x_2,\dots,x_m)$ *characteristics* is defined by *syntactic rules of WFF construction*. Any syntactic rule defines a sign lying in the basis of expression symbols ordering. Each characteristic conjunctively links formula symbols, but formula meaning is not the result of Boolean operation. In this case WFF do not contain variables and are suppositions [3].

Then,

$$z_1=<K_1,V_1,X_1,A_1,C_1> \qquad (2)$$

Any WFF of $z_1$ system is presentation and according to axiom 1 – true formula, i.e. axiom.

Let $F$ be WFF set of $z_1$ system, then:

$$A_1=F=T_1 \quad (3)$$

$C_1$ set of true FCR deduction rules in $z_1$ consists of a just one rule of separation (modus ponens). This rule is used for matching any axiom with its number, i.e. for axioms numbering:

$$F_i\,(x_i\,)\to i \qquad (4)$$

where $i$ is the number of axiom. This is bijective mapping.

Then, the following rules will be added to $V_1$ set:

3. $i$ is WFF.
4. if a certain subset of obtained axiom numbers can be ordered according to definite $x_j$ characteristic, then

$$F_j(x_j\,)=(i_1,i_2,\ldots,i_k) \qquad (5)$$

is WFF.

5. Recursively, if $F_i(x_i\,)$ is WFF, the expression

$$F_1(x_1)\to F_2(x_2)\to\ldots\to F_n(x_n) \qquad (6)$$

is also WFF.

Let us define the features of $z_1$ system.

*Feature 1.* $F$ set is WFF, consequently $A_1$ set of $z_1$ system axioms are recursively countable and finite.

In fact, is $z_1$ is an actual system, then memory resources and time of its existence are finite, therefore $F$ and $A_1$ are finite. As all WFF are numbered according to the rules of WFF construction $F$ and $A_1$ are recursively countable.

*Feature 2* will be stated as the following theorem.

*Theorem 1.* $Z_1$ formal system is consistent and complete.

Theorem proving. A set of $z_1$ system expressions to which syntactic rules of $V_1$ and $X_1$ sets cannot be applied make up $R_1$ set. In this system false expressions do not have structural value (semantic meaning), so they are not numbered and in the course of further data processing are not used. Then, $R_1=\varnothing$. As, according to (3) $T_1=F$, $T_1$ is finite and recursively numerable set, so $T_1\cap R_1=\varnothing$, i.e. the use of $V_1$ и $X_1$ rules distinguish the given sets. Therefore, $z_1$ is consistent system.

As there exists efficient algorithm of deduction with the use of WFF construction rules, so any WFF in $z_1$ system can be proved, then $z_1$ is complete system.

It is evident, that in the process of $z_1$ system functioning there can always be found $F'$, that is the extension of $F$: $F\subset F'$. As $F'$ is a WFF set, it is also finite and recursively numerable. Consequently, $z_1$ system is complete and consistent even within this extension.

It can be concluded that $z_1$, system which has 1 and 2 features makes it possible to solve the problem of *image identification*.

In a general view, $\mathcal{F}(q_n)$ image (presentation) of $q_n$ objects that belongs to CE, in $z_1$ system is WFF:

$$\begin{array}{ccccccccc} q_n & \rightarrow & \left[\begin{array}{c} F_1(x_1)\\ F_2(x_1)\\ \vdots \\ F_k(x_1)\end{array}\right] & \rightarrow & \left[\begin{array}{c} F_1(x_2)\\ F_2(x_2)\\ \vdots \\ F_k(x_2)\end{array}\right] & \rightarrow \cdots \rightarrow & F_n(x_n) & & \\ & & \| & & \| & & \| & & \\ & & \mathcal{F}_1(x_1) & \rightarrow & \mathcal{F}_2(x_2) & \rightarrow \cdots \rightarrow & \mathcal{F}_n(x_n) & = & \mathcal{F}(q_n) \end{array} \qquad (7)$$

Then $q_n \Leftrightarrow i$ mapping, where '$\Leftrightarrow$' is the sign that means 'if and only if', exists, in the case if $\mathcal{F}_n(x_n) \rightarrow i \equiv \mathcal{F}(q_n) \rightarrow i$, where '$\equiv$' is the sign of equivalence. Then, $q_n$ object is said to be identified in $z_1$ system by $i$ name.

To make $z_1$ systems *classify objects*, it is necessary to add the following rule of $A_1$ subdividing into classes (subclasses) to $V_1$ set of rules:

6. *Classification rule*. If $\mathcal{F}(q_1) \rightarrow i_1$ and $\mathcal{F}(q_2) \rightarrow i_2$ are WFF, where $i_1$ and $i_2$ are axiom numbers which can be semantically interpreted as the names of $q_1$ and $q_2$ ($q_1 \neq q_2$) objects in $z_1$ system ($q_1 \in Q$, $q_2 \in Q$, where $Q$ is the class of objects) and exists

$$\mathcal{F}(q_1) \cap \mathcal{F}(q_2) = \mathcal{F}(Q) \qquad (8)$$

then $\mathcal{F}(Q)$ is WFF.

Then, $\mathcal{F}(Q) \rightarrow i$ implication defines the name of $i$ class of $Q$ objects.

However, in order to make $q_1$ and $q_2$ objects belong to $Q$ class it is essential but insufficient to meet the conditions (8).

In fact, there exist a set of different WFF intersections, but not all of them distinguish axiom classes constructively, i.e. are essential and sufficient features of classes.

Belonging of objects and, consequently presentations (images) in $z_1$ to either class can be defined only as a result of system learning "with" or "without the teacher".

$Z_1$ system learning "without the teacher" or system self-teaching is based on the meeting the following condition:

*Condition 1.* If there exists a set of CE objects $\aleph = (q_1, q_2, \ldots, q_n)$, that are successively perceived by $z_1$ system ($\aleph$ set will be called *learning sample*) and there exists a set of pair-wise intersection $\mathcal{F}(q_i) \cap \mathcal{F}(q_j) = \mathcal{F}(q^*)$; $i \neq j$, then subset of objects $\aleph^* \subseteq \aleph$ belongs to $Q$ class if and only if there is

$$\mathcal{F}(Q)=max/\mathcal{F}(q^*)/ \qquad (9)$$

for all $q_i$, $q_j \in Q$.

According to this condition two images belong to the same class if their WFF have the same substructures of symbol sequences and these sequences have maximum length.

*Definition 4.* WFF of (9) type will be called *class WFF, class axiom* or *Conc(Q) concept.*

This axiom must be stable, i.e. keep the order and number of element in the process of learning. Class axioms are added to general list of axioms and participate in the process of learning equally with other axioms. During the other cycle of learning a new more constructive class axiom (axiom that separates subsets of axioms more efficiently) can be formed.

Evidently, learning without the teacher has more iterations and depends on the capability of learning sample and sequence of its elements.

Learning "with the teacher" is more complex and efficient according to the number of iterations. ‘Teacher’ is considered to be an external system with respect to $z_1$ system. This system will be designated as $z_2$.

Prior to consideration this system it is necessary to note the following general feature of $z_1$ type systems.

*Feature 3.* If two systems $z_1$ and $z_1'$ have learning samples $\aleph$ and $\aleph'$ ($\aleph \neq \aleph'$) respectively, then WFF (axioms) numbers of similar presentations in these systems will not coincide.

This evident feature sets the following basic requirement to $z_2$ system.

*Requirement 1.* Whatever the numbers of axioms of analogous presentations $\mathcal{F}(q)$ and $\mathcal{F}'(q)$ in systems $z_1$ and $z_1'$ are, these axioms must implicatively be brought in correspondence with the similar symbols of $K_2$ alphabet $z_2$ system.

If:

$$\mathcal{F}(q)\rightarrow i \text{ и } \mathcal{F}'(q)\rightarrow j,\ i\neq j,$$

then

$$\mathcal{F}(q)\rightarrow i\rightarrow b_1;$$
$$\mathcal{F}'(q)\rightarrow j\rightarrow b_1, \qquad (10)$$

where $b_1 \in K_2$.

Thus, $z_2$ system will be over-system or ‘teaching system’ for any system of типа $z_1$ type.

As symbols of $K_2$ alphabet can be brought in one-to-one correspondence to the numbers of $z_1$ system axioms by only $z_2$ system directly in the process of their development using learning procedure with teacher, then $z_2$ system must be FIS subsystem provided that $K_2$ alphabet is identical for all FIS.

Interaction of different FIS and their subsystems is shown in Fig.1, where $\mathcal{K}$ is bijective mapping of $K_2$ alphabets of $z_2$ and $z_2'$, systems, which formalizes corresponding communicative function between FIS-1 and FIS-2. In Fig. 1 sign

"$\leftrightarrow$" means bijective mapping, i.e. if $f(A)=K_2$, then $f^{-1}(K_2)=A$, or $A\leftrightarrow K_2$. To implement communicative function $\mathcal{K}$ it is necessary that communicative subsystems of $z_3$ type with $K_3$ alphabet exists in each FIS.

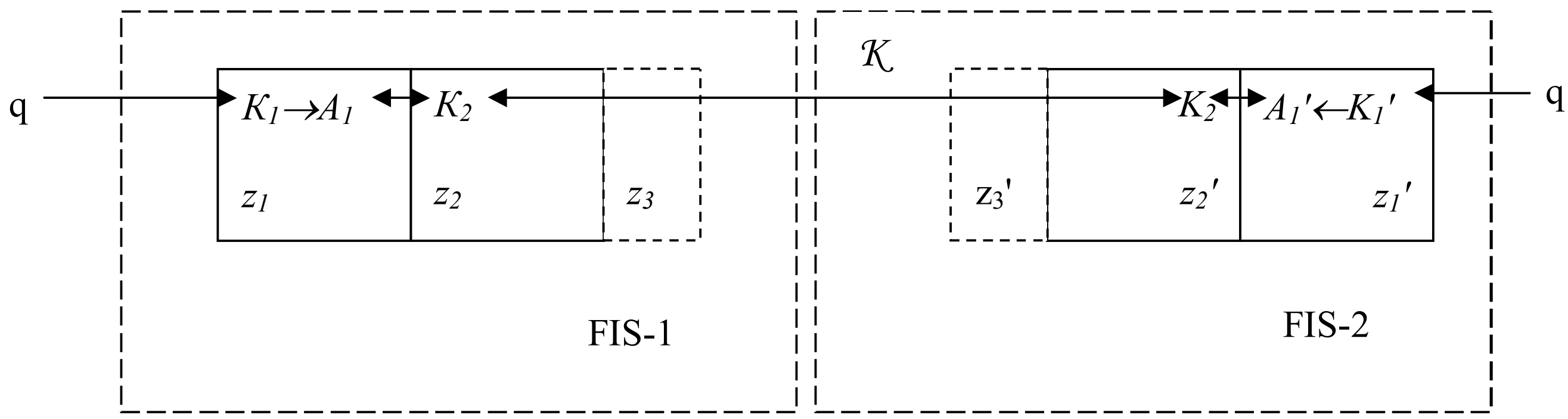

Fig. 1. FIS interaction while implementing communicative function $\mathcal{K}$

*Definition5.* Thus, in $z_2$ system repeated presentations of perceived objects are formed in $K_2$ alphabet which will be called *representations*, $z_2$ system being called *representation subsystem or system* (RS).

The main feature of $z_1$ system 'with teacher' (by $z_2$ system) is meeting the following condition.

*Condition 2.* If there exists a set of presentations $\mathcal{F}(q_1), \mathcal{F}(q_2), \ldots, \mathcal{F}(q_m)$ and $f^{-1}$ mapping matching any presentation with the same $b_1$ letter of $K_2$ alphabet

$$f^{-1}(\mathcal{F}(q_1), \mathcal{F}(q_2), \ldots, \mathcal{F}(q_m)) = b_1 \qquad (11)$$

which can be presented as the diagram:

$$\left.\begin{array}{l} \mathcal{F}(q_1) \leftarrow \\ \mathcal{F}(q_2) \leftarrow \\ \vdots \\ \mathcal{F}(q_m) \leftarrow \end{array}\right( \quad b_1$$

then there exists

$$\mathcal{F}(Q) = min/\mathcal{F}(q^*)/ \qquad (12)$$

where $\mathcal{F}(Q)$ is an axiom of $Q$ class; $min/\mathcal{F}(q^*)/$ is the minimum length of symbols sequence in a set of pair-wise intersections $\mathcal{F}(q_i)\cap\mathcal{F}(q_j)=\mathcal{F}(q^*)$; $i\neq j$.

Therefore, there exists $f$ mapping:

$$f(\mathcal{F}(Q)) = b_1,$$

then

$$\mathcal{F}(Q) \leftrightarrow b_1 \qquad (13)$$

Thus, obligatory class features are sorted out and class axiom is formed even during the first iteration while 'with teacher' learning, i.e. external (with respect to $z_1$), 'forced' class separation. It makes this type of learning efficient.

Impossibility to develop efficient procedure of classification in $z_1$ system without $z_2$ system is not the only limitation of $z_1$. The main limitation of this system is its feature *4 which* can be formulated as the following theorem.

*Theorem 2.* System $z_1$ is closed loop relative to $T_1$ set.

Theorem proving. According to expression (3), $T_1$ is finite and recursively numerable set. Let us suppose that there exists recursively numerable infinite truth set $\mathcal{T}$ and $T_1 \subset \mathcal{T}$. Then there exists open set $U = \mathcal{T} \setminus T_1$, that is $T_1$ addition to $\mathcal{T}$ and, therefore, $T_1$ is closed set.

$Z_1$ system is set by tuple (2) and according to (3) $A_1 = T_1$ only $T_1$ set from the given tuple is constructively augmented set.

*Definition 6. Constructive augmenting* of $T_1$ set of $z_1$ system is considered to be its extension as a result of new presentations development in the process of system functioning.

Thus, $z_1'$ extension of $z_1$ system exists if and only if when there exists $T_1' \supset T_1$ extension; the latter makes it possible to state that $z_1$ is closed relative to $T_1$.

The most important FIS feature is formulating new axioms as a result of proving or disproving generated hypotheses in the process of system functioning. Theorem 2 shows that $z_1$ system cannot generate, prove or disprove hypotheses. This function is accomplished by $z_2$ system.

*Definition 7.* The hypothesis $h_i \in H$ will be considered to be WFF of $z_2$ system that was formed as a result of performing $\mathcal{K}$ communicative function or under the influence of other FIS subsystems, it being unknown whether they are true or not.

Let us consider the features of RC $z_2$.

In this system any generated WFF $h_i$ can refer either to $T_2$ set or to $R_2$ set. Thus, in $z_2$ system unlike in $z_1$ system the set of false expressions is not empty $R_2 \neq \varnothing$.

Being a formal system, $z_2$ must be described by the following tuple:

$$z_2 = \langle K_2, V_2, X_2, A_2, C_2 \rangle \qquad (14)$$

where:

*$K_2$ alphabet* is presented by a finite set of symbols that are *semantic determiners* (SD) or $A_1$ axiom numbers.

$$K_2 = (b_1, b_2, \ldots, b_m) \qquad (15)$$

where $b_i$ is $K_2$ alphabet symbols.

The following feature of $z_2$ system is valid.

*Feature 5.* $K_2$ alphabet is closed relative to $A_1$ set.

Proceeding from *requirement 1*, *condition 2 , and theorem 2*, this feature proving is evident.

Thus, $z_1$ system limits $z_2$ system.

Construction rules of $V_2$ set WFF will be determined as following:

1. Any $b_i$ symbol of $K_2$ alphabet is WFF.
2. If

$$G_i(x_i)=(b_1, b_2, \ldots, b_n) \qquad (16)$$

is the sequence of $K_2$ alphabet symbols ordered according to $x_i \in X_2$ characteristic, then $G_i(x_i)$ is WFF. As $G_i(x_i)$ does not contain variables, then it is a sentence.

3. If $G_i(x_i)$ is implicatively matched to a definite number $i$ of a given sentence

$$G_i(x_i) \rightarrow i \qquad (17)$$

then $i$ is WFF.

4. If a certain subset of WFF numbers can be ordered according to $x_j \in X_2$ characteristic, then the expression

$$G_j(x_j)=(i_1, i_2, \ldots, i_k) \quad (18)$$

is WFF as well.

5. Recursively: if $G_i(x_i)$ is WFF, then the expression

$$G_1(x_1) \rightarrow G_2(x_2) \rightarrow \ldots \rightarrow G_n(x_n) \qquad (19)$$

is WFF too.

So, in $z_2$ system there appears a new type of WFF – these are $h_i \in H$ hypotheses.

Any $h_i$ hypothesis is either a sentence of $G_i(x_i)$ type or $P_i(x_i)$ predicate in which some or all elements of the formula are replaced by $y_i$ variables.

*Definition 8.* $Y_i$ variable will be considered to be an undefined element that takes the *i*-place in the sequence of WFF symbols.

Predicate truth or falsity depends upon the interpretation of variables. Thus, variables specify the degree of uncertainty while deducing axioms in $z_2$.

*Definition 9. N*-place $P_i(x_i)$ predicate in $z_2$ will be considered to be a linear-ordered structure – WFF which is composed of $n$ elements of WFF type and/or variables and ordered according to $x_i \in X_2$ characteristic, where $P_i$ is a predicate name or *predicate symbol*.

*Definition 10.* $P_i \in K_2$ predicate symbol is either 1) SD of a process (action) in which predicate elements take part, or 2) SD of predicate elements belonging to a certain set (subset) of axioms, or 3) SD of interconnection of predicate elements that belong to different sets (subsets) of axioms.

*Definition 11.* The place in the predicate structure that can be taken by its element will be called an *actant*.

An actant semantically defines the role of an element in a predicate. The quantity of actants defines the predicate dimension.

Thus, the predicate will be considered to be rather linguistic than mathematic construction. Predicate symbol makes the stable nucleus of a formula and linguistically defined the "predicate". In fact, the type of the process does not change (is invariant) relative to the objects that participate in this process.

Then, in $V_2$ set the following rule can be added:

6. $P_i(x_i)=(\mathcal{G},\Upsilon)$ is WFF, where $\mathcal{G}$ is the set of WFF in predicate structure that are the symbols of $K_2$ alphabet, and $\Upsilon$ is the set of variables in the predicate structure that take place of corresponding actants.

In the expression (20) 'empty' predicate structure is shown, where the sign '_' defines the place of a certain actant.

$$P_i(_,_,\ldots,_) \qquad (20)$$

In a general case $P_i(x_i)$ predicate structure is different from $G_i(x_i)$ sentence structure; in the predicate not all actants are filled in, free actants can be filled in only by $K_2$ alphabet symbols according to $x_i$ syntactic rule. From the formal point of view an empty actant is interpreted as a variable. $G_i(x_i)$ sentence will have the same actant structure as $P_i(x_i)$ predicate, where $G_i$ name is analogous to predicate symbol $P_i$.

It is evident that any proved predicate or sentence, if it is absent in $A_2$ set, supplement this set.

Then $A_2$ set will constructively augment as a result of proving $h_i$ hypotheses.

Theorems in $z_2$ are deduced with the use of the following rules which form $C_2$ set:

1. use of separation rule (4);
2. identification of equivalence:

$$h_i=G_i(x_i) \Leftrightarrow f^{-1}(G_i(x_i))= \mathcal{F}(q_i) \qquad (21)$$

3. substitution of $y_i$ variable in $P_i(x_i)$ by $K_2$ alphabet symbols.

Each expressions of $G_i(x_i)$ or $P_i(x_i)$ type is constructed on the basis of $x_i \in X_2$ *syntactic rule.*

If $x_i \in X_1$ are given by basic, fundamental, and derived characteristics, then $x_i \in X_2$ are determined by linguistic constructions formed in the process of $z_2$ system learning 'with the teacher', i.e. with the participation of another FIS.

The essence of syntactic rules of $X_2$ set is in forming types and sequence of actants for every predicate or sentence.

Then,

$$/X_2/=/P/=/G/ \quad (22)$$

For the English language types of actants and their sequence in a predicate or sentence are determined by prepositions of learning sample in the process of performing $\mathcal{K}$ communicative function.

Let us analyze false expressions in $z_2$.

It is evident, that concept 'false' can be applied only to $h_i$ hypothesis, i.e. to the expression formed in $z_2$ under the influence of other subsystems of the given FIS or other FIS.

If $h_i=G_i(x_i)$, then WFF will be determined as true on applying rule (21). This rule of deduction actually checks whether there exists $\mathcal{F}(q_i)$ presentation that corresponds to representation expression $G_i(x_i)$. In the case when this mapping does not exist or it cannot be deduced from the axioms that are present in $z_1$, then $G_i(x_i)$ is a false expression.

If $h_i=P_i(x_i)$, then it is possible to define whether a predicate is true on applying the following derivation scheme:

*Derivation scheme 1.*

1. $f^{-1}(P_i)=F_i$ is the identification of a process (action) that corresponds to a predicate symbol.
2. $f^{-1}(\mathcal{G})=\mathcal{F}_i$ is the identification of objects that correspond to WFF present in predicate actants.
3. If $\mathcal{F}_i$ is a part of $F_i$ locus, then $P_i(x_i)$ is true provided $\mathcal{Y}=f(\mathcal{F}_j)$, where $\mathcal{F}_j$ is also a part of $F_i$ locus (replacement of $\mathcal{Y}$ variables of predicate at presentations that correspond to actant types).

*Locus* will be considered to be a distinguished subset of presentations that are directly interconnected with the examined presentation that is in the centre of the locus. In a general case, $F_{i-1}$, $F_{i+1}$ in $F_{i-1}\rightarrow F_i\rightarrow F_{i+1}$ structure, i.e. directly interconnected with $F_i$, form $F_i$ *locus*.

After performing 1-3 steps of the given scheme, $P_i(x_i)$ can be considered as *actual representation*, i.e. $P_i(x_i)$ is *actually true*.

After performing 1 and 2 steps of the given scheme, i.e. without substitution variables by symbols, $P_i(x_i)$ can be considered as *conditionally true*. In this case a predicate can either true or false depending on certain interpretations of variables.

Then it is possible to formulate the following statement.

*Statement 1.* The absence of variable interpretation in $P_i(x_i)$ predicate result in the lack of grounds for solving $h_i=P_i(x_i)$ hypothesis (reference to $T$ or $R$).

Let us define the following criteria for predicate falsity:

1. Predicate symbol or actant symbols are not identified in $z_1$.
2. Presentations of actant symbols are not a part of the locus of predicate symbol presentation.
3. Predicate is actually true if actants have other meanings. For example, the predicate that corresponds to the linguistic construction – 'Michael left home at 14.00' contradicts the presentation that corresponds to the linguistic construction ‘Michael Brown left for work at 8.30’ according to the actant of time of action.

4. If mapping $f^{1}(P_i(x_i))=\mathcal{F}_i(x_i)$ hereditarily contradicts presentations present in $z_1$. For example, the predicate that corresponds to the linguistic construction ‘A human being lives eternally’ contradicts the presentation that corresponds to the linguistic construction ‘Alan Turing died on June 7, 1954’. Here ‘Alan Turing’ is an example of a human being; the event limited by time – ‘died’ – contradicts the event unlimited by time – ‘lives’.

Let us introduce negation symbol ‘¬’ (‘not’). This symbol in the hypothesis $\neg P_i(x_i)$ or $\neg G_i(x_i)$ is the direction to check whether corresponding presentation is absent in $z_1$.

In this case the derivation scheme will be the following:

*Derivation scheme 2.*

1. $f^{-1}(P_i)=F_i$;
2. $f^{-1}(\mathcal{G})=\mathcal{F}_i$;
3. If $\mathcal{F}_i$ is not a part of $F_i$ locus, then $\neg P_i(x_i)$ is true;
4. If $\mathcal{F}_i$ is a part of $F_i$ locus, then $\neg P_i(x_i)$ is true if $\Upsilon=f(\mathcal{F}_j)$, where $\mathcal{F}_j$ is not a part of $F_i$ locus.

If symbol ‘¬’ connects an actant symbol in the predicate but not a predicate symbol, for example $P_i(x_i)=(b_1,b_2,-,-,\ldots,\neg b_k,\ldots,b_n)$, them the derivation scheme will be the following:

*Derivation scheme 3.*

1. $f^{-1}(P_i)=F_i$;
2. $f^{-1}(b_1,b_2,\ldots, b_k,\ldots,b_n)=(F_1,F_2,\ldots,F_k,\ldots,F_n)$;
3. If $f^{-1}(b_k)=F_k$ presentation is not a part of $F_i$ locus, then $P_i(x_i)$ is true if when there are corresponding interpretations $\Upsilon=f(\mathcal{F}_j)$, where $\mathcal{F}_j$ is a part of $F_i$ locus.

Let us consider *Feature 6* of $z_2$ system that is stated as a following theorem.

*Theorem 3.* Isolated formal system $z_2$ is unsolvable.

Theorem proving. Let us suppose that $z_2$ system is isolated relative to $z_1$, i.e. these systems are not interconnected. Then derivation schemes 1-3 are inapplicable. In this case, $h_i$ hypotheses can be defined as true or false only on the basis of the given set $A_2$, rules of WFF construction $V_2$ and $X_2$, rules of derivation 1 and 3 from $C_2$ set. Then efficient procedure of theorem derivation will be absent in $z_2$, as only logically valid WFF can be true in $z_2$. Any other hypothesis of $G_i(x_i)$ type can be obtained from $P_i(x_i)$ as a result of random substitution of $y_i$ variables by $K_2$ alphabet symbols. To define whether this expression is true is possible only after adding this expression directly to $T_2$ set using teaching FIS. If this teaching procedure is absent, then according to Gödel’s incompleteness theorem [4], it is always possible to develop true WFF which are unprovable. Besides, it is possible to develop WFF that cannot be proved or disproved (for example, known logical paradoxes).

Thus isolated formal system $z_2$ is unsolvable. Besides, it is always possible to find such proved FCR that will be disproved as a result of $A_2$ set expansion during learning 'with the teacher'. Thus, isolated formal system $z_2$ is also contradictory and incomplete.

According to the given theorem isolated system $z_2$ have the features of formal system on the basis of classical calculus of predicates limited by the fact that $A_2$ set can be augmented only as a result of realization of communicative function $\mathcal{K}$.

Feature 7 of $z_2$ system results from theorem 3.

*Feature 7.* $K_2$ alphabet of isolated formal system $z_2$ is inexpressible relative to objects of perception.

In fact, $K_2$ alphabet is inexpressible because it is not possible to perform one-to-one mapping of a certain object of perception $q_i$ in its SD $b_i \in K_2$, i.e. $q_i \Leftrightarrow b_i$ interconnection is absent.

As SD corresponds only to the presentation of the highest level of generality (7):

$$q_i \Leftrightarrow (\mathcal{F}(q_i)= \mathcal{F}_1(x_1) \rightarrow \mathcal{F}_2(x_2) \rightarrow \ldots \rightarrow \mathcal{F}_n(x_n)) \qquad (23)$$

and

$$\mathcal{F}_n(x_n) \leftrightarrow i \leftrightarrow b_i,\ b_i \in K_2,$$

then the absence of $f^{-1}(b_i)=\mathcal{F}_n(x_n)$ mapping is equivalent to the absence of, $q_i \Leftrightarrow b_i$ interconnection.

This evident feature has an important consequence.

*Consequence 1.* Isolated representation systems are inexpressible relative to the objects of perception.

This means that to construct FIS it is necessary to have both presentation and representation systems.

The interconnection of systems $z_1$ and $z_2$ makes it possible to prove the following theorem (*feature 8*).

*Theorem 4.* $Z_2$ system is solvable relative to $z_1$ system.

Theorem proving. According to feature 5, $K_2$ alphabet is closed relative to $A_1$ set which, in its turn according to theorem 2, can be constructively augmented as a result of developing new presentations. The interconnection of systems $z_1$ and $z_2$ makes it possible to apply derivation schemes 1-3, that makes it possible to prove or disprove any hypotheses $h_i$ on the basis of their check in $z_1$. That indicates an efficient algorithm capable (using finite number of steps for any WFF) to define whether this WFF is a theorem or not (whether it belongs to the set of true formulae $T_2$ or to a set of false formulae $R_2$), that makes it possible to state that $z_2$ system is solvable relative to $z_1$ system.

It is evident, that even in this case $z_2$ system is incomplete and inconsistent.

One of the most important features of $z_2$ system is the following:

*Feature 9.* $A_2$ set is formed: 1) as a result of $A_1$ mapping in $K_2$ alphabet symbols; 2) as a result of proving $h_i$ hypotheses; 3) as a result of $A_2$ expansion while $\mathcal{K}$ communicative function is being realized.

Thus, a great number of external FIS as well as subsystems of the given FIS can participate in the development of $A_2$

*Definition 12*. Let us name this feature the *representation diversity* of FIS.

In this feature $A_1$ set is a *presentation base*.

Representation diversity shows the capability of FIS to generate and check (prove or disprove) hypotheses.

Thus, the following principle of FIS development can be formulated:

*The principle of representation diversity in FIS development.* The wider the presentation base (more capable $A_1$), the more subsystems of this FIS and external systems participate in the process of $A_2$ development, the higher representation diversity of FIS.

Considering this principle, it is possible to make the following conclusions:

1. FIS must have at least two subsystems – presentation and representation ones.

2. To create FIS that has the same representation diversity as a human intelligence it is necessary to simulate all subsystems or perception, presentation, and representation of a human being, to teach FIS using powerful learning samples and a great number of learning systems.

## 2.2 Principles of FIS construction and functioning

All the above arguments are based on the premise of the existence of *effective resolution algorithm*. We show that this efficient algorithm exists, and define its characteristics.

This paper considers the general approach to the problem of construction of presentations of static single contour images that consist of straight line segments on a surface.

The basis of FIS construction is neural elements – detectors of perception, presentation and representation of income images. The set of similar detectors that define (construct) the presentation of the images of the same class pattern recognition make up *zones* of detectors. Detectors making up such zone will be termed *zone detectors*. The zones of detectors are FIS memory. Since zones are made at different steps and stages of data processing, FIS memory is distributed among all its structures. Data in FIS is processed at each stage due to operation of detectors of specific structure, i.e. *processors*. Generally, while processing data upstream, processors perceive output signals of zone detectors which are placed in lower hierarchical zones and excite certain zone detectors that are placed in higher hierarchical zones. Processor and its interrelated zones make up *module* of data processing step. All modules operate according to the same principle and structurally intersect at the zone level. Data processing stage consists of several steps. Neuron structure that comprises a great number of step modules makes up a module of data processing stage.

Signal that is formed at the output of any zone detector is the response to certain sequence of signals that are fed to its inputs from detectors of a given module and/or other modules and subsystems of FIS. Zone detector that has an output signal will be termed *excited detector*.

Depending on the functions that different detectors perform while processing data at the initial level of FIS PSS, the following types of zone detectors will be marked out:

1. Detectors which respond to structural elements of input images and images in general. These detectors will be termed *structural detectors*. Structural detectors integrate identification and classification features of the structure of image elements, single image, a great number of images that are perceived simultaneously (scenes), etc. Structural detectors responses will be brought in correspondence with the elements of $K_1$ alphabet of PSS. The simplest or primitive structural element while considering CE images is the point of image contour. Structural detectors responses are the result of implicative convolution of input sequence of signals that consists of responses of both structural and characteristic detectors. Structural detectors are the central elements of presentation synthesis that function as so-called 'granny cells' [2]. Sets of structural detectors of the same class pattern recognition make up *structural zones*.

2. Detectors which respond to certain classification and identification features (characteristics) of structural elements of images and images in general. These detectors will be termed *characteristic detectors*. Responses of characteristic detectors are the result of implicative convolution of input sequence of signals that comprises only responses of characteristic detectors of the same type. These detectors are designed to create $X_1$ set of syntax rules of WFF construction (to form ordered chains of elements of PSS $K_1$ alphabet), as well as to classify and identify. Responses of certain types of characteristic detectors make classifying (zone or sub-zone) responses which, as well as structural detectors responses, correspond to $K_1$ alphabet elements. These detectors define classes (sub-classes) of presentations. The sets of characteristic detectors of the same type make up *characteristic zones*.

Thus, the main objective of zone detectors is memorizing certain input signal sequences and responding at the output.

The sequence of input signals of zone detectors is constructed by processing detectors or processors on basis of output signal selection that are fed from zone detectors of lower hierarchical level and sorting these signals in accordance with certain algorithm. Processors are universal detectors in the sense that they do not memorize input signal sequences and have dynamic data interrelation with any detectors either of lower or higher hierarchical zones of a certain module.

With parallel spatial perceiving the input images of FIS PS at a fixed period of time, construction of linearly ordered set of detectors responses is a mechanism of ordered synthesis of presentations.

Linearly ordered set of responses of structural or characteristic detectors for CE model presents certain geometrical structure of image contour that consists of, for example, straight and/or curve line segments.

The basis of constructing any linearly ordered structure (sequence) is the *selecting* the points (elements) of the start (*pb*) and *searching* (defining) of the points of the end (*pe*) of the order.

The formalization of *pb-pe* selecting procedure in FIS is one of the most important problem, as its solving within the higher nervous activity of a human being is based not only on the work of the system of involuntary attention that uses short (direct) control communications (e.g. pre-motor theory of attention) but also on the action of integrated system of voluntary (selective) attention that uses long (transitive, descending) control connections (e.g. super-motor theory of attention) [5] and connected with the emotional system as well as with other regulatory systems of the brain.

In the considered FIS only visual system of perception is modulated. The following items can be referred to the mechanisms of involuntary attention of visual system: a) the mechanism of eye focusing at a certain stimulus, i.e. object of perception, intended for segmenting the environment of visual perception; b) saccadic eye movements, a function of which being segmenting the object structure that is placed in the zones of the utmost eye sensitivity (fovea); c) the mechanism of work of peripheral 'watchdogs', i.e. neurons that are placed out of the fovea zone and actively react to stimuli movement or change of brightness or colour.

One of the simplest models of mechanism of attention in FIS may be procedure of scanning PS field of perception in a certain direction and object 'capture' when critical points of the first kind (the first detected points of image contour) are detected. These points belong to a certain polygon hypothetically outlined around the image contour and define the convexity of image structure.

### 2.2.1 Zone detectors

Let us consider the essence of zone detectors responses. Any $d_i$ detector construct at its output a unique signal, i.e. the response to input signal sequence from certain detectors placed, for example, on a lower hierarchical level of data processing. The uniqueness of any detector is defined by its place in FIS structure, i.e. by its address (number). Thus, every detector responds to its excitation by 'placing' its address at the output.

Let us suppose that any zone detector forms only one output signal, i.e. each detector is 'assigned' to a certain presentation structure (substructure).

Let us suppose that at the initial time of FIS functioning detectors are not interconnected with information, that means that each detector 'does not know' which input signal sequence formed by the processor it must respond to, forming at its output a signal that identifies its address. Signal that provides such connection is the control signal 'capturing' the considered detector that is fed from a particular detector that controls the zone, i.e. zone identifier (ZI), that corresponds to the step or stage of data processing. ZI finds free, non-initialized detectors in a definite zone and initialized them assigning free detector to a new sequence.

Zone brain structure is well known, there exist detailed brain maps that reproduce functional differentiation of individual structures of neurons. In the same way, a zone that comprises a number of detectors is reserved for every symbol of FIS $K_1$ alphabet. Any zone is *flexible*; it can stretch or shrink while FIS functioning, so it can change its structure. In particular, zone flexibility becomes apparent when 'capturing' free detectors that do not belong to the considered zone; when its own resources are exhausted, or when transferring 'redundant' detectors to other zones.

This definition of zone is based on the supposition that every symbol of $K_1$ alphabet that belongs to certain category of presentations (category of recognition) is assigned to an individual detector. If we suppose that the detector can make a lot of identifying responses, it becomes evident that zone definition will be quite different. In this case, new detectors will appear in the zone due to the limitations of computational resources (memory) of each detector. For simplicity, let us keep in consideration the first definition of zone.

Processor reserves zone 'capturing' it competitively. Zone 'capture' means initialization of its 'key' detector – ZI.

ZI defines zone address – zone response of detectors that belong to the zone, controls zone detectors, i.e. 'captures' detectors or transfer them to other zones, structures the zone, i.e. orders detectors inside the zone, creates sub-zones and their identifiers. If the zone 'is captured' (initialized), then at its ZI output an output signal is generated, i.e. zone address in response to certain sequence of input signals.

ZI forms structural or characteristic cluster, i.e. serves for classifying presentations, any detector that belongs to the zone identifying particular presentation (image sample).

Thus, ZI must only respond to the sequence of input signals that corresponds to stable features (characteristics) of the classification. Therefore, ZI of any zone only responds to *the sequence of signals fed from characteristic detectors*. This sequence is standard for a particular zone, i.e. *Conc*($Q_i$) *concept* of category of $Q_i$ image presentations. The concept is formed by an identifier (it is selected from the general sequence of input signals) and is corrected (changed) in the process of FIS learning either "with" or "without a teacher".

Inside the zone detectors can be grouped in sub-zones that correspond to stable sub-structures of presentations. Each sub-zone must have its own identifier (SZI) responding to the corresponding sub-structure of the concept. In the majority of zones there exits the hierarchy of sub-zones, i.e. the structure of interrelations between ZI, SZI and zone detectors inside every zone realizes *genus-species relations*.

Thus, ZI and SZI, unlike static zone detectors, are flexible with regard to input data sequence. They are designed to create, memorize, and modify concepts in the process of learning and serve for structuring (address defining) of their zone detectors. These identifiers in the process of data processing function as data flows switches.

It is evident that the structure of *Conc*($Q_i$) concept that defines CE $Q_i$ image presentation is stable (invariant) with regard to different changes of perceived image.

For CE images, considering all the limitations suggested by this paper, such changes can include:

a. affine transformations of image contour (turning or changing the place of the image in the perception coordinate system (PCS), change of image scale);

b. deformation changes of image contour:

- of the 1st genus, i.e. addition or subtraction of separate elements of the contour which does not result in the change of presentations class;

- of the 2nd genus, i.e. change of the dimensions of separate elements of the contour which does not result in the change of presentations class.

Concept invariance with regard to affine transformations and deformation changes of the 2nd genus is achieved by abstracting from unstable characteristics which can change in the process of perception of different image samples. These characteristics are deduced from the structure of the concept during the process of learning.

Concept invariance with regard to deformation changes of the 1st genus is achieved by optimizing the sequence of responses of structural detectors while constructing inductive structures [6].

Thus, complex response at the output of the *i*'th *detector* $d_i$ that is placed in the *j*'th sub-zone of the *k*'th zone is defined by the following elements: classification constituent, i.e. zone address and sub-zone address within the zone; identification constituent, i.e. the address of the considered detector within the zone (sub-zone) $d_{kji}$.

### 2.2.2 The levels of excitation of zone detectors

Let us specify three levels of zone detectors excitation which define three different states: *actual level* (actual state), i.e. the level of maximum excitation; *latent level* (the state of pre-choice) and *the level of residual excitation* (quiescent state), i.e. the level of minimum. Actual level and the level of residual excitation have a number of values, i.e. degrees of excitation.

So, at the actual level a detector has several degrees (values) of excitation depending on the chain length, i.e. the quantity of sequential elements in the order formed by the processor that agree with the concept. The more the characteristic signals at the output of the processor agree with the concept, the higher the level of actual excitation of ZI and corresponding zone detector is. If this sequence agrees with a certain zone concept *Conc*($Q_i$) completely, then ZI of this zone and corresponding zone detector will have maximum degree of actual excitation.

Let us designate the *i*'th detector of the *k*'th zone that is in the state of actual excitation *a,* as $d^{a}_{ki}$; in the state of latent excitation *l* as $d^{l}_{ki}$; in the state of residual excitation *r* as $d^{r}_{ki}$.

During certain period of time $\Delta t_{sel}$ the processor selects from a number of zone detectors with different level of excitation only detectors that have actual level of excitation. At every level of data processing there can be a number of characteristic and structural detectors with actual level of excitation. Detectors in the actual state can be considered to be at the 'top' of data processing. The processor successively orders the responses of these detectors with regard to the response of the desired detector that is *pb* up to the moment when detector *pe* is detected, if the given detector is not specified by the other module or sub-system.

Any detector can be in the actual state during the period of time $\Delta t_{sel}$ necessary for its selection by the processor. After this the level of detector excitation automatically decreases up to the latent level. The detector is converted into the actual state by zone identifier during ascending data flow processing or by the *control signal of excitation* of descending data flow.

Let us suppose that at the period of time $t_0 \in \Delta t_{sel}$ on the *i*'th level of data processing two alternative detectors $d_{ki}$ and $d_{k+1,i}$ are actually excited simultaneously. So, the processor is tasked to select one detector on the *i+1*'th level.

Let us define that if during $\Delta t_{sel}$ the detector $d^{a}_{ki}$ , that has actual level of excitation *a*, is influenced by the *control signal of inhibition* $h^{-}$ from the other detector of the given level of data processing that is in the actual state ($h^{-} \rightarrow d^{a}_{ki}$) and $h^{-} > a$ ($h^{-}$ – dominating level of signal), then the level of $d^{a}_{ki}$ detector excitation decreases up to the latent *l* ($a - h^{-} = l$). If $h^{-} < a$ (*a* – dominating level of signal), then $a - h^{-} = a$, i.e. this detector remains in the actual state during the whole period of time $\Delta t_{sel}$.

Let us assume that the length of coincident chain in the structure of presentation $Q_i$ that is formed by the processor and Conc*(k)* is greater than in Conc*(k+1)*. Then, the level of actual excitation of $d^{a1}_{ki}$ detector is higher than of detector $d^{a2}_{k+1,i}$ ($a1 > a2$). Consequently, the level of inhibition signal $h^{-1} = a1$ that influences $d^{a2}_{k+1,i}$ detector is higher than the level of inhibition signal $h^{-2} = a2$ that influences $d^{a1}_{ki}$ detector.

Then:

$$a1 - h^{-2} = a; \; a2 - h^{-1} = l;$$

$$(h^{-1} \rightarrow d^{a2}_{k+1,i}) = d^{l}_{k+1,i}; \; (h^{-2} \rightarrow d^{a1}_{ki}) = d^{a1}_{ki}. \qquad (24).$$

Thus, the actual level of excitation and its degree are the criteria for the processor in selecting a definite zone detector for its subsequent processing; and the mechanism of inhibition eliminates alternative variants, i.e. directly simulates the selection process.

This process will be termed the *competition between the levels of excitation* or α- *competition*.

Detectors with the latent level of excitation that have direct interconnection with the *central detector* in its actual state make up its *locus*.

So, if $d^{a}_{ki}$ detector is in the actual state, then all detectors of its *k'*th zone and ZI of this zone belong to its locus within horizontal connections (*horizontal locus*); all the detectors of the previous level of data processing, the sequence of which resulted in actual excitation $d^{a}_{ki}$ ,belong to the locus within vertical connections (*vertical locus*).

If the detector is converted into actual state, then all the detector of its locus are converted into the latent state *l* either automatically when $\Delta t_{sel}$ is over (within vertical connections), or with the help of signal of inhibition/excitation ($h^{-}/h$), that is fed within horizontal connections. This signal generated, for example, by $d_{ki}^{a}$ signal is interpreted by all the detectors of the locus that are in the actual state as the signal of inhibition $h^{-}$, and by the detectors of the locus that are in the state of residual excitation *r* as the signal of excitation *h* up to the level *l*. Thus, in all FIS modules a 'wave' of actual excitation and subsequent inhibition takes place in the process of transition between the levels of ascending data flow processing.

Locus detectors that are in the state of latent excitation provide the opportunity of their alternative selection by the module for ordering. The transition between the locus detectors makes it possible to 'complete' or 'restore' the order; this is the basis of decision-making process in FIS.

Therefore, the mechanism of creating locus detectors and transitions between them is the mechanism of hypotheses formation in the result of 'restoring' incomplete presentations or *synthesis* of alternative presentations.

Thus, in a general view the decision-making process in FIS can be considered as a process of linear ordering detectors of a certain level of data processing between the detectors of *pb-pe* order that can be specified by the system of attention, RS, or other modules or sub-systems of FIS.

If the detector that is in the latent state does not belong to the locus, then its excitation decreases to the residual level *r*. The residual level of excitation is the main factor of formation of both long-term memory and operational memory. On this level there are also several degrees of excitations. The degree of residual excitation depends on the frequency of excitation, the period of excitation (the time of image exposure), the time that has passed since the time of the last excitation and emotional component of excitation. Singly excited detector has minimum level of the residual excitation. If the detector is not re-excited for a certain time, then the degree of its residual excitation decreases up to zero. A detector with zero degree of residual excitation is a free detector and it can be 'captured' by a new sample of $Q_i$ image. This is the way of realization of one of the mechanisms of *memory compression*, i.e. '*forgetting*'. This mechanism also defines the flexibility of zones.

The structure of zone detector interconnections with the other detectors of FIS PSS is given in Fig. 2.

This figure shows the following types of signals:

$d_i$ – these are the sequence of input signals that was formed by the processor of data processing step;

$e_i$ – these are output signals of excitation/inhibition to the detectors of the vertical locus;

$h'/h^{-\prime}$ – these are input signals of excitation/inhibition from the detectors of the horizontal locus;

$h/h^{-}$ - these are the output signals of excitation/inhibition to the detectors of the horizontal locus;

$d_{i+1}$ – this is a detector output signal;
$e_{i+1}$ – this is an input signal of excitation/inhibition within the horizontal locus;
$z$ – this is a signal of 'capturing'/initializing a detector of ZI (SZI).

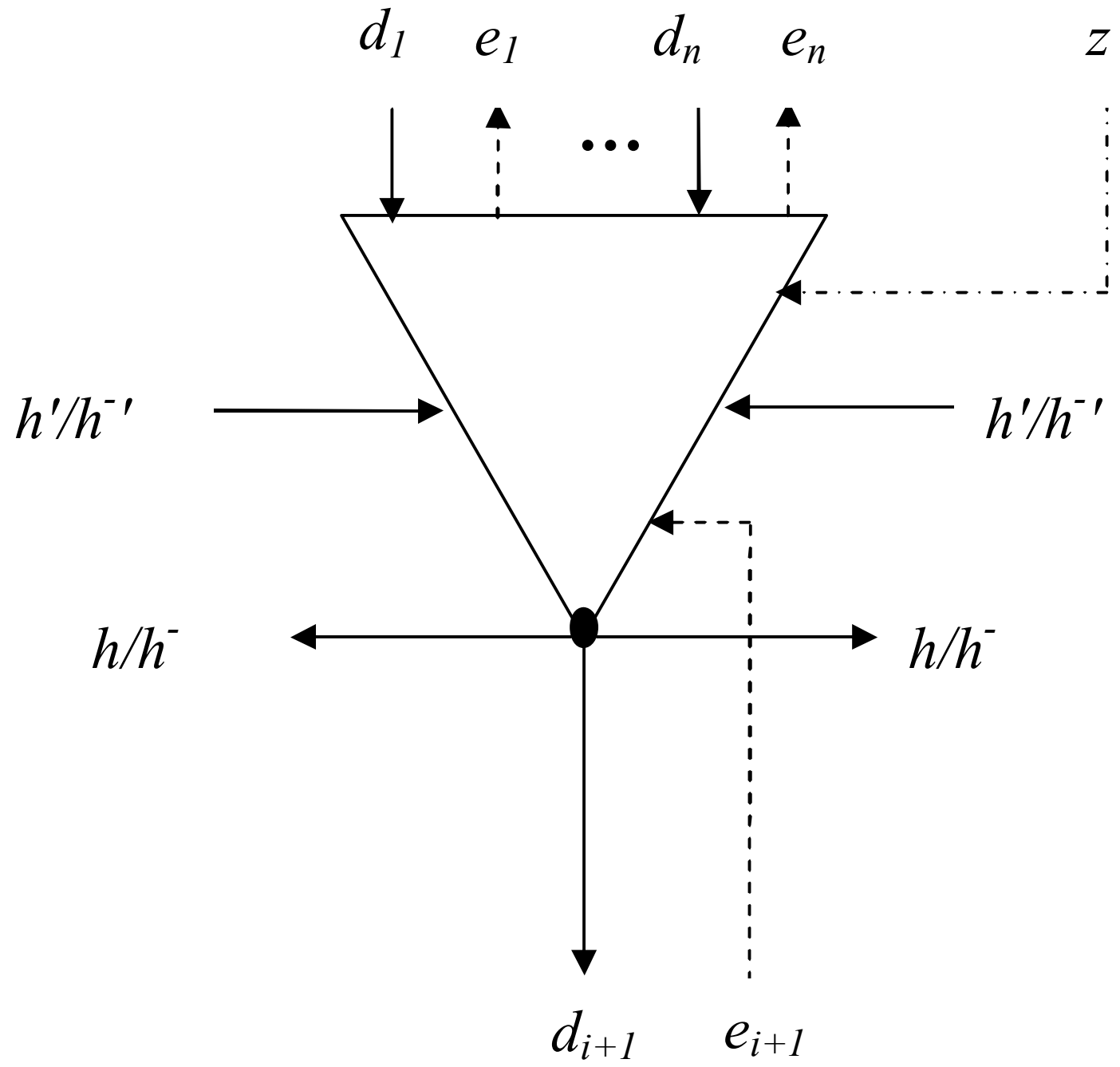

Fig. 2. The structure of zone detector interconnections

**2.2.3 Processors**

These neural elements in data processing act as data flows filters.

Each stage of data processing in FIS is connected with a definite internal presentation or representation system of coordinates. For example, at the stage of perception this is the perceptive coordinate system (PCS) that defines the arrangement of stimuli (primitive structural elements of contour image, i.e. dots) in the field of perception; at the stage of making presentations of individual images and scenes this is the orientation coordinate system (OCS); at the stage of making presentations of elements of the world spatial model this is the egocentric coordinate system (ECS); at the stage of making presentations of events extended on time this is temporal coordinate system (TCS); at the stage of making representations this semantic coordinate system (SCS). Each coordinate system has its own axes, i.e. a number of characteristics.

At the end of each stage only one structural detector responds to each input sequence of detector responses. This response is a 'dot' in the coordinate system of the following stage. A number of data processing stages in FIS depends on a number of its perceptive sub-systems, attention, motor or other sub-systems as well as on a number of characteristics in each of them.

The response of each structural detector is the result of implicative convolution of a number of responses of structural and characteristic detectors excited while data processing at every stage.

Each stage consists of several steps; at each of these steps responses of structural and/or characteristic detectors that were obtained during the previous step are ordered and generalizing responses of the detectors of corresponding type are generated.

To order detectors responses during every step of making presentations we will mark out individual types of processors.

Processors that order the responses of only structural detectors will be termed *structural processors*; processors that order the responses of only characteristic detectors will be termed *characteristic processors*; processors that order the responses both structural and characteristic detectors at the same time will be termed *structural-and-characteristic detectors*.

Detectors that are in the actual state are selected by a certain processor and their responses are transferred (projected) to the strictly specified inputs of this processor. Each processor input is connected with the corresponding element of its internal linear structure. We will term these processor elements *modes*. There are *structural modes* and *characteristic modes* of a certain characteristic.

A sub-set consisting of the modes of the same type will be termed correspondingly *structural* or *characteristic modal group*.

Ordered modal group or a set of ordered modal groups in the processor will be termed the *vector of modes*. The vector of modes forms 'a cut' of sets of different characteristics in a definite direction.

A block diagram of *Pr* processor is given in Fig. 3.

This figure sows: *A* – the unit of actual input response; *B* – the unit of generating output signals of excitation/inhibition *e*; $d_1$ - $d_n$ – the modes of structural modal group; $d_{n+1}$ - $d_m$ – the modes of characteristic modal group.

For CE model, considering all the limitations suggested by this paper, the following types of characteristics for PCS and OCS will be defined:

1. *Basic characteristics* ($x_1$).

Basic characteristics define the exes of presentation coordinate systems, i.e. a number of vectors of characteristic or structural modes ordering.

A basic characteristic that defines the direction of construction of any mode vector will be termed *structural basic characteristic* (SBC).

For CE model SBC defines the direction of execution of the image contour with regard the initial point of ordering (the point of 'contour capturing') clockwise (left-to-right) or anticlockwise (right-to-left). For simplicity, we will consider only one value of this characteristic – this is the direction of ordering clockwise. The value of this characteristic is built into the operating algorithm of any processor.

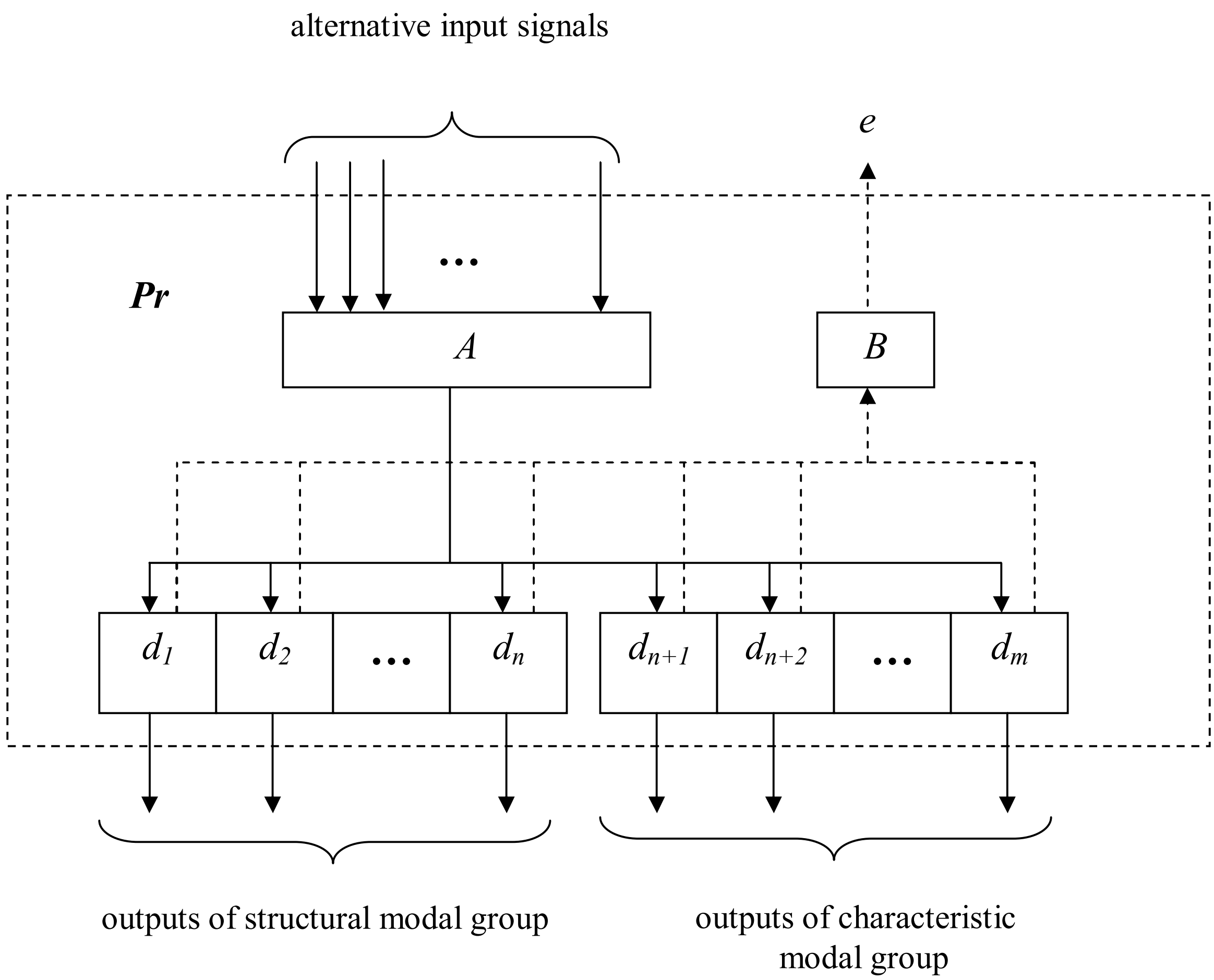

Fig.3. Block diagram of the processor

Besides SBC, basic characteristics connected with the corresponding presentation coordinate system are defined at every level of data processing. Basic characteristics can be binary or n-ary and are specified structurally (algorithmically) in the process of FIS designing.

A number of vectors of orienting image contour segments or vectors of development (change of characteristics) of image contour sub-structure (structure) are referred to the basic OCS characteristic (the example is given in Fig.4). According to the results of neurobiological researches, in the striate cortex of cerebrum there are the columns of neurons that response to the change of stimuli (segments) orientation with the step no less that 1 degree [2];

The transition to a new coordinates system and, consequently, to new basic characteristics defines the stage of data processing.

2. *Fundamental characteristics* ($x_2$).

Fundamental characteristics are used to:

a. create space-time model of external (with regard to FIS) environment and its structuring;
b. define the stages of data processing.

These characteristics define, for example, the following sequence of ascending data flow processing in FIS for CE model: presentation of elementary images, isolated images, scenes and events stretched in time, etc;

c. identify and classify images.

Time and space characteristics are fundamental ones for FIS PSS.

Time characteristics are different time periods connected with data processing at different stages of image perception and presentation as well as the characteristics of time interleaving of perception and time scaling of events that is connected with relative perception (past, present, or future), and perception of astronomical time.

Space characteristics are the ones that define the arrangement of elements of image structure or the whole image in different FIS presentation coordinate systems. Space characteristics work as the basis of creating presentation spatial model of environment in ECS. In particular, in biological systems these characteristics define retina-shaped structure of striate (visual) cortex of cerebrum.

These characteristics are present in mode vectors at all stages of data processing and they do not depend on the structure of image. Scaling one or another fundamental characteristic in the corresponding coordinate system is the peculiarity of every data processing stage.

3. *Derived characteristics* ($x_3$).

Values of these characteristics show the number, quantity, relation or type of changing fundamental, basic or other derived characteristics.

Derived characteristics are generated by the processor as a result of *sequential comparison* of similar modes values that are ordered within the mode vector. Following the comparison, they can either *coincide* or *differ*. Correspondingly, these characteristics will be termed *the characteristics of coincidence* or *difference.* New derived characteristics are formed during each step of data processing.

Derived characteristics can be:

a. *Quantitative,* that have multiple values.

These characteristics include, for example, the length of a segment, a number of segments or angles, angular value, etc. Sequential coincidence of values of structural or characteristic modes results in the quantitative characteristic, i.e. the length of continuous chain of coincident modes. One characteristic may have two different values that are sequential in a mode vector; if so, it can be a basis for further characteristic decomposition, i.e. for creating new derived characteristics that define new ‘quality surge’ of two structural elements that are compared. This ‘surge’ value and direction (vector) are defined by characteristic processor of the following step. This processor will be termed *characteristic decomposition processor*.

Thus, the difference of quantitative characteristics values results in the qualitative characteristics that have binary values at the following step to data processing.

b. *Qualitative*, that have binary or n-ary (multiple) values.

Binary qualitative characteristics are, as a rule, classifying (zonal) component of difference quantitative characteristic. Identifying component of this characteristic will be the value of this difference.

Binary characteristics include, for example: increase or decrease of qualitative characteristics values, right (clockwise) or left (anticlockwise) change of orientation direction of a segment or sub-structure development, contour closure or break, similar or different ordering of certain characteristics within the mode vector that defines, for example, whether the whole image contour or its elements are convex or concave, as well as the order with monotonous (successive) or discrete (jumping) change of characteristics, etc.

If ordering binary qualitative characteristics of image elements does not result in forming binary values of qualitative characteristics of the whole image (the are no similar concepts), then n-ary qualitative characteristics of the whole image are formed.

Thus, *decomposition of characteristics lasts up to the ordered sequence of qualitative binary characteristics of the same type, i.e. n-ary qualitative characteristics of the whole image that is perceived.*

Derived characteristics are formed:

- for image substructures;
- for the whole image.

Let us consider the main steps that exist at every stage of data processing examining the stage of making the presentation of a separate angle. Let the angle is shaped with two segments *a* and *b* of the lengths *n* and *m* (*n*>*m*) and the directions of orientation of OCS *f* and *g* (Fig.4).

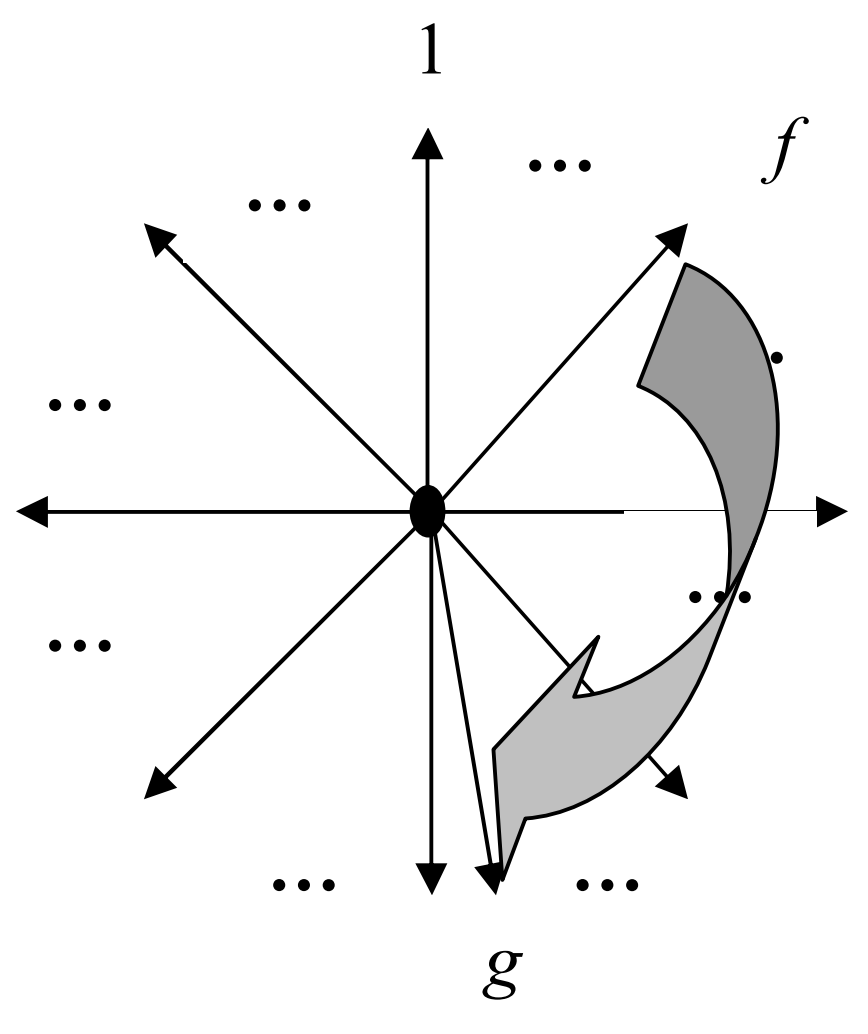

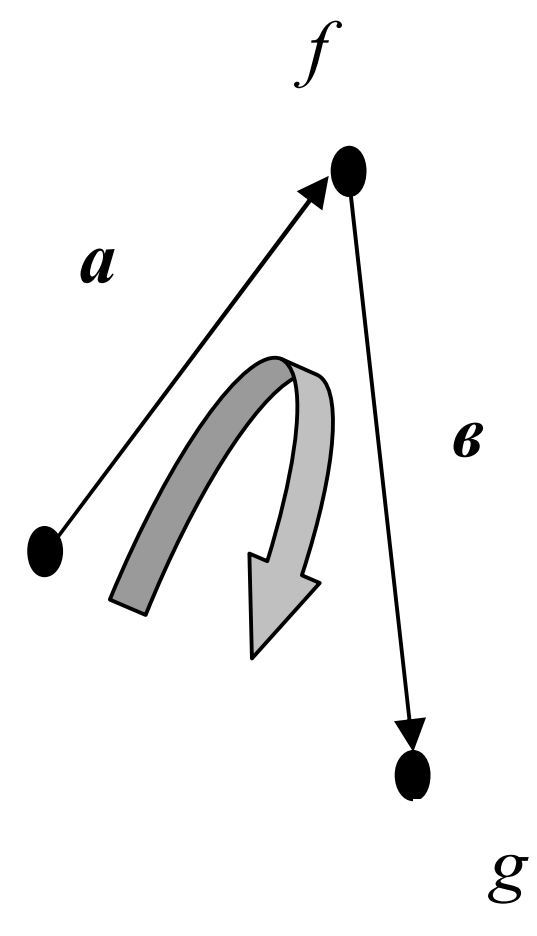

Fig. 4. a) OCS, b) angular characteristics in OCS

For simplicity, we will not consider the fundamental space characteristic of the segments arrangement in PCS.

Let us assume that the responses of presentation structural detectors of these segments have already been formed at the previous stage. Let these responses be correspondingly $d_{nfi}$ and $d_{mgi}$, where: *i* is classifying address of the structural zone of segments presentation; *f* and *g* are classifying addresses of orientation sub-zones; *n* and *m* are identifying characteristics that define the addresses of the corresponding detectors in the sub-zones.

Every stage module begins with the step of structural and characteristic integration, i.e. 'assembly' of modes vectors from responses of structural or characteristic obtained at the previous stage. This step is performed by one *integration structural processor* and several *integration characteristic processors*.

At the ***first step*** (Fig. 5.) *integration structural processor A* forms a mode vector on basis of results of $\alpha$- competition detectors of the *i*'th structural zone. The vector comprising $d_{nfi}$ and $d_{mgi}$ structural modes is ordered according to SBC (the direction of contour execution), where $d_{nfi}$ is *pb,* and $d_{mgi}$ is *pe* of order.

This mode vector is identifying component of the response of structural detector of the whole angle image presentation, so it is completely projected onto the *identifying header* of the mode vector this is formed by the *processor of structural and characteristic synthesis D* at the third step of this stage. Thus, the integration structural processor is peculiar as there are no zone detectors excited by it. It is designed only for 'assembling' the vector of structural modes, for initializing the process of characteristic decomposition, for forming structural identifying features of the presentation being made, as well as for structural transitions between presentation levels (stages) while reverse propagation of control signals for locus detector excitation.

While modes vector constructing, the integration structural processor *A* performs characteristic decomposition of structural detectors responses as a result of sequential shaping reverse control signals, i.e. 'echo' signals, which along vertical connections convert $d_n$, $d_f$, $d_i$ detectors that belong to the vertical locus $d_{nfi}$ detector, as well as $d_m$, $d_g$, $d_i$ detectors that belong to the vertical locus of $d_{mgi}$ detector, into the actual state. These detectors project (transmit) their responses to the corresponding *integration characteristic processors* $B_1$, $B_2$, $B_3$.

Sequential shaping 'echo' signals by the integration structural processor defines the time sequence of excitation of detectors of the corresponding locus, which, in its turn, results in the sequential interpretation ('filling' with the responses of excited detectors) of the modal group of one or another integration characteristic processor.

The process of sequential in time 'capturing' the modes in the processors modal groups by the detectors that are in the actual state will be termed *time competition* or *t- competition*.

Thus, in this example $B_1$, $B_2$, $B_3$ integration characteristic processors form modes vectors that consist of two modes of each characteristic ordered according to the time of excitation of corresponding detectors which, in this case, is like SBC vector. While constructing modes vector each integration characteristic processor *compares* the modes values of only one characteristic.

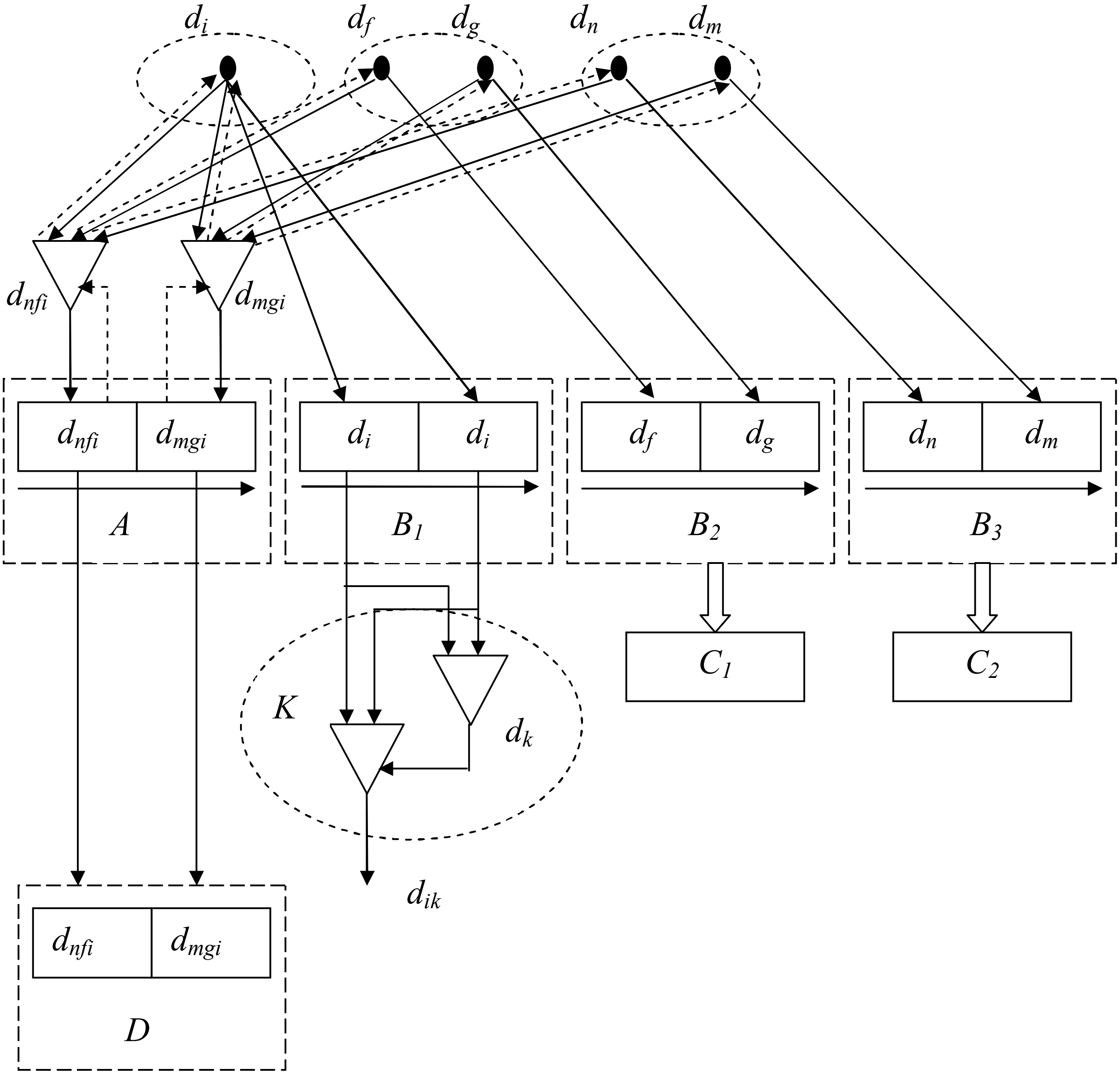

Fig. 5. Step # 1. Structural and characteristic integration

So, $B_1$ processor compares the response of the same $d_i$ detector that have *identical values* obtained at the successive periods of time $t_i$ and $t_{i+1}$. The coincidence of modes values result in the excitation of corresponding ZI and creation of actual response of $d_{ik}$ detector in characteristic zone *k* of this processor. This response defines a number of segments (elements, sub-structures) that creates the considered image.

$B_2$ processor compares the responses of $d_f$ and $d_g$ detectors that have *different values* and also obtained at the successive periods of time $t_i$ and $t_{i+1}$. Difference of responses indicates the necessity of further characteristic decomposition, i.e. transition to $C_1$ processor of the next step of data processing. Similarly, $B_3$ processor compares the responses of $d_n$ and $d_m$ detectors that have different values and directs the control to $C_2$ processor.

At the ***second step*** (Fig. 6) characteristic decomposition takes place, that is obtaining new derived characteristics on basis of ordering *difference values of characteristic modes* for sub-structures of the perceived image (if any) or the whole image. At this step data is processed by *characteristic decomposition processors*.

The peculiarity of the processor of this type is the fact that its first mode – *pb* is the projection of the first mode compared in the processor of the previous step, and the latter one, i.e. *pe*, is the projection of the second compared modes. The points of beginning and end of the vector that is being formed are defined in this way. It is necessary to 'restore' the missing chain of responses of the detectors of a certain zone in the mode vector that is being formed. To do this the processor shapes the control *signal of excitation*, i.e. 'echo' signal that converts the zone detector into actual state, its response being reflected in *pb* mode; $d_f$ detector will be used for $C_1$ processor, and $d_n$ detector will be used for $C_2$ processor. These detectors are the centres of horizontal locuses in corresponding zones. The centers of horizontal locuses send control signals of excitation along the horizontal connections in the direction of *pb-pe* vector, converting the nearest neighbouring detectors $d_{f+1}$ and $d_{n+1}$ of its horizontal locus into actual state. The responses of these detectors are projected into corresponding modes of $C_1$ and $C_2$ processors.

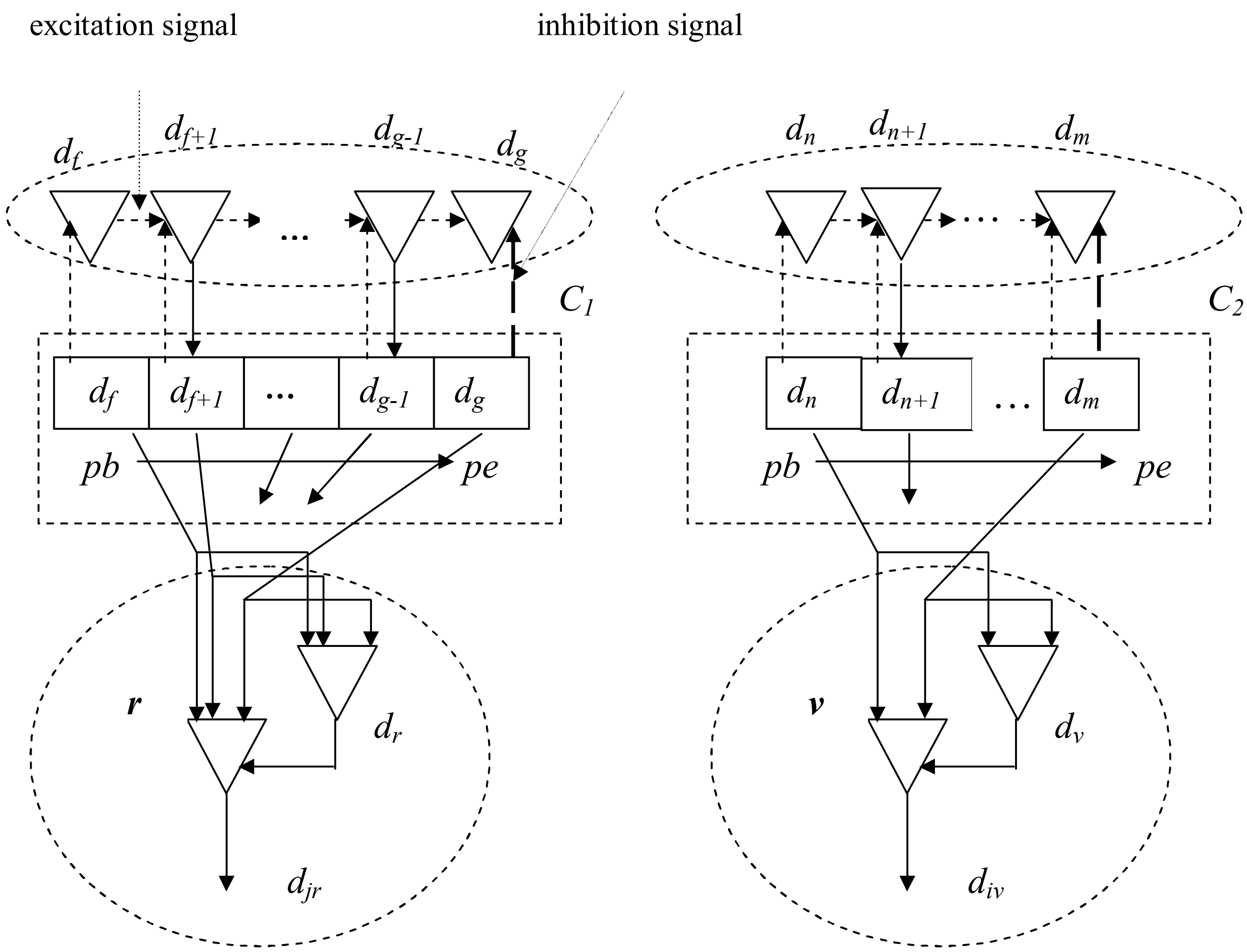

Fig. 6. Step # 2. Characteristic decomposition

Then, shaping the sequence of ‘echo’ signals $C_1$ and $C_2$ processors initialize the successive transition along the detectors of horizontal locuses as well as simultaneously project their responses into the modes of these processors up to the moment of $d_g$ and $d_m$, detectors excitation, their responses defining *pe* points in the corresponding vectors of modes. When these detectors are converted into the actual state, the further propagation of control signals of excitation along the horizontal connections stops as the processors generate control *signals of inhibition* that are fed along the vertical control connections. So signals of inhibition that are fed along the vertical locuses break the process of further transitions between zone detectors in horizontal locuses.

The vectors of modes are formed from the detectors values that monotonously change (are continuous in zones) on a basis of *t*-competition. Classifying (zone) component of the vector of modes for $C_1$ processor will be qualitative binary characteristic of the direction of orientation change of the segments that shape the angle, for the considered example this is *r* direction. For $C_2$ processor this will be a qualitative binary characteristic of the change of these segments length, for the considered example the value of this characteristic is *v* reduction.

Identifying component of the modes vector formed by $C_1$ processor will be the value of the change of orientation directions of the segments that shape the angle, i.e. the angular value *j* in the OCS values that is defined as a number of orientation directions within *pb-pe* gap.

Thus, the vector of modes that is formed by $C_1$ processor specifies *r* characteristic zone as a result of excitation of $d_r$ ZI that, in its turn, forms the actual response of $d_{jr}$ detector that belong to this zone.

Identifying component of the vector of modes that was formed by $C_2$ processor will be the value of difference *i* of the length of the segments that shape the angle.

Then, the vector of modes that was formed by $C_2$ processor specifies the characteristic zone *v* as a result of excitation of $d_v$ ZI and defines the actual response of $d_{iv}$ detector of this zone.

At the ***third step*** (Fig. 7) the responses of the structural detector of the whole image presentation are formed (in the considered example this is an angle) as a result of the operation of the processor of *structural and characteristic synthesis D*.

The peculiarity of the operation of the processor of this type is forming the vector of modes as a result of *t*-competition of actual responses of structural and characteristic detectors obtained at the previous steps of the stage. So, the sequence of structural detectors (the vector of modes) that was formed by *A* processor at the first step at the moment of time $t_1$ is projected into the structural modal group (identifying header) of the modes vector that is being formed *D* processor and consists of the responses of $d_{nfi}$ and $d_{mgi}$ detectors. Then, at the moment of time $t_k$ the actual response of $d_{ik}$ detector excited as a result of $B_1$ processor operation is projected into the characteristic modal group of *D* processor. At the moment of time $t_n$ actual responses of $d_{jr}$ and $d_{iv}$ detectors excited as a result of operation of $C_1$ and $C_2$ processors are projected into this modal group.

Characteristic modal group on the formed vector of modes defines the classifying component of the presentation of the considered image. Interpreted (filled) characteristic modal group is projected into ZI of the structural zone of *D* processor. For the considered stage of data processing the processor of structural and characteristic synthesis *D* has only one structural zone of *i+1* presentation (at the following steps processor of this type will have a set of alternative structural zones of the presentation).

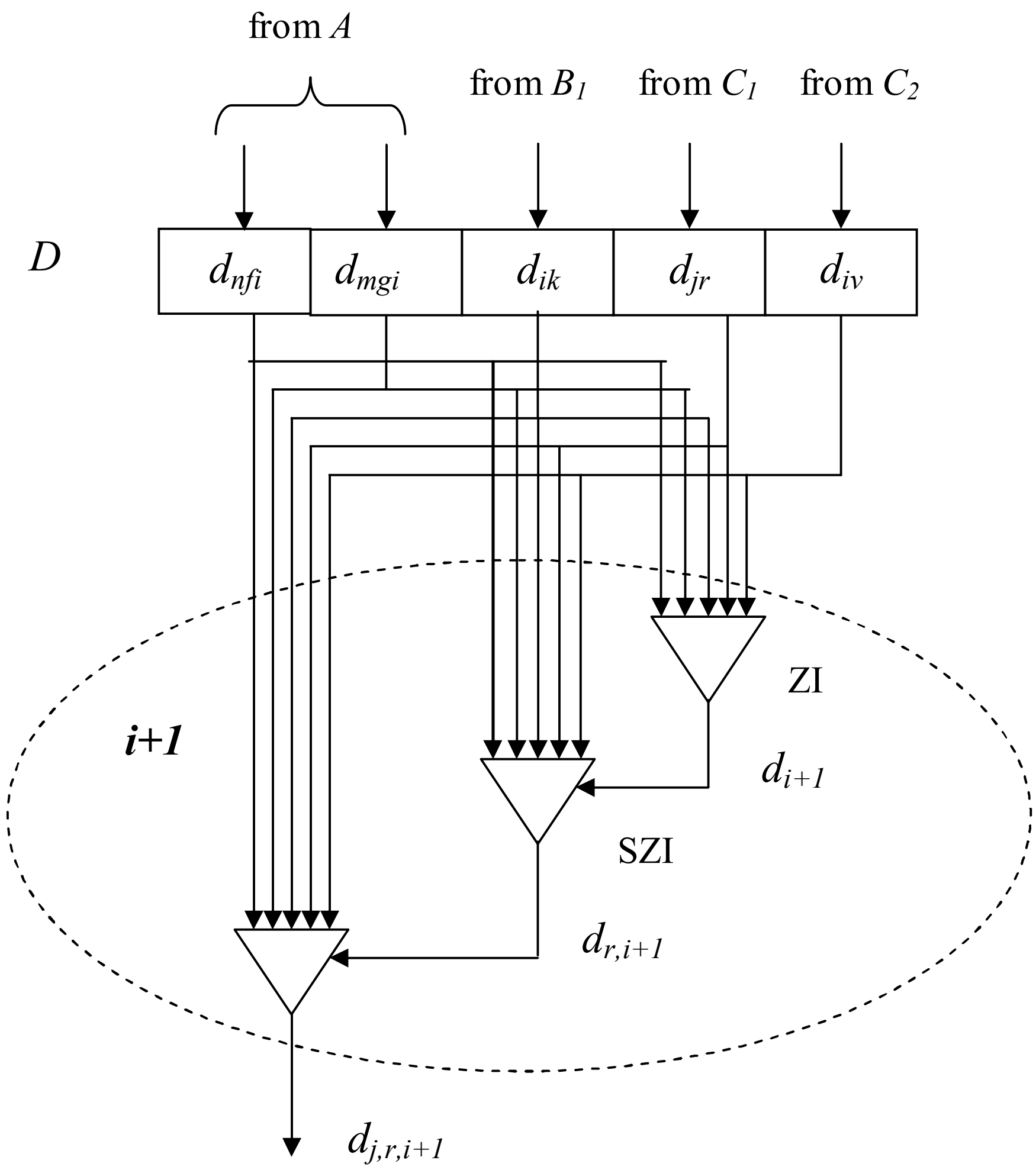

Fig. 7. Step # 3. Structural and characteristic synthesis

ZI compares the values of modes fed into its inputs with the concept (reference sequence) formed and stored by ZI in the process of learning 'with the teacher' in the process of self-learning. For the considered example, as there are no alternative zones of presentation, the procedure of self-learning is like the procedure of learning 'with the teacher', i.e. the sequence of signals at the ZI inputs is either identical to the concept or modifies it in the direction of minimizing a number of stable features of recognition that are expressed as a the sequence of values of characteristic modal group.

Let us assume that the concept of ZI ($d_{i+1}$ detector) is already formed and comprises one characteristic mode with $d_{ik}$ value. This value is the sequence of responses fed to the input of this detector. $d_{i+1}$ detector is converted into the actual state and using its output response to enable projecting output sequence of *D* processor signals into the inputs of SZI ($d_{r,i+1}$ and $d_{l,i+1}$ detectors), where: $r$ is the right direction of changing segments orientation that shape the angle, and $l$ is the left one. The concepts of these detectors correspondingly comprise $d_r$ and $d_l$ values. In the input sequence of signals there is $d_{jr}$ value. So, $d_{r,i+1}$ detector is converted into the actual state. It, in its turn, enables projecting output sequence of $D$ processor signals into the inputs of detectors of sub-zone.

Let us assume that in the sub-zone there is no detector with formed response to this input sequence, i.e. there are no 'echo' signals from sub-zone detectors converted into the actual state with the complete coincidence of the sequence of signals at their inputs with the sequence stored earlier. Then SZI of $d_{r,i+1}$ forms the control signal 'capturing' free detector $d_{j,r,i+1}$. According to this signal $d_{j,r,i+1}$ detector stores the sequence of signals at its input and forms the output response. This output response is presentation structural response of a particular perceived sample image (instant) of 'angle' class.

However, the idea of 'angle' is $a_i$ symbol of $K_2$ alphabet of FIS RS. In this example this symbol is *semantic determiner* (SD) of the class of perceived images with respect to detectors of FIS PSS. In the process of FIS learning 'with the teacher' this SD is implicatively brought in correspondence with ZI, i.e. $d_{i+1}$ detector, so $d_{i+1} \rightarrow a_i$ connection is set. Thus, if $d_{i+1}$ detector of FIS PSS is excited, then $a_i$ detector of FIS RS is also excited. In this case $d_{j,r,i+1}$ zone structural detector does not have direct connection with RS detectors of FIS, and consequently, its excitation will not convert any PS detector into the actual state.

The general diagram of steps interconnection at any stage of ascending data flow processing is given in Fig. 8, where: $A$ are integration processors, $B$ are characteristic decomposition processors, $C$ are processors of structural and characteristic synthesis.

In a general case, structural zone detectors of the $i$'th and $i+1$ 'th stages are interconnected as '*a part to the whole'*.

Let us formulate the following statements that define the construction of the whole FIS PSS.

*Statement 2.* Any stage of ascending data flow processing consists of 3 steps: the step of structural and characteristic integration, the step of characteristic decomposition, and the step of structural and characteristic synthesis.

*Statement 3.* The processors of each step of any stage operate according to the same algorithm.

*Statement 4.* Algorithms forming responses of zone detectors, responses of ZI and SZI detectors are universal for all FIS modules.

For the considered CE model, the sequence of stages of ascending data flow processing defines the sequence of the levels of presentation of perceived images: segments, angles, individual images shaped by the angles, a set of simultaneously perceived images, i.e. scenes. Data at these stages is processed according to the characteristics of PCS and OCS. The resulting presentation of the listed stages is the response of the detector that defines 'a point' in ECS.

It is necessary to mention that retina-shaped spatial structure of arrangement and interconnection of processors and zone detectors is maintained only at these levels. However, ZI and SZI arrangement in the zones does not confirm to retina-shaped relations.

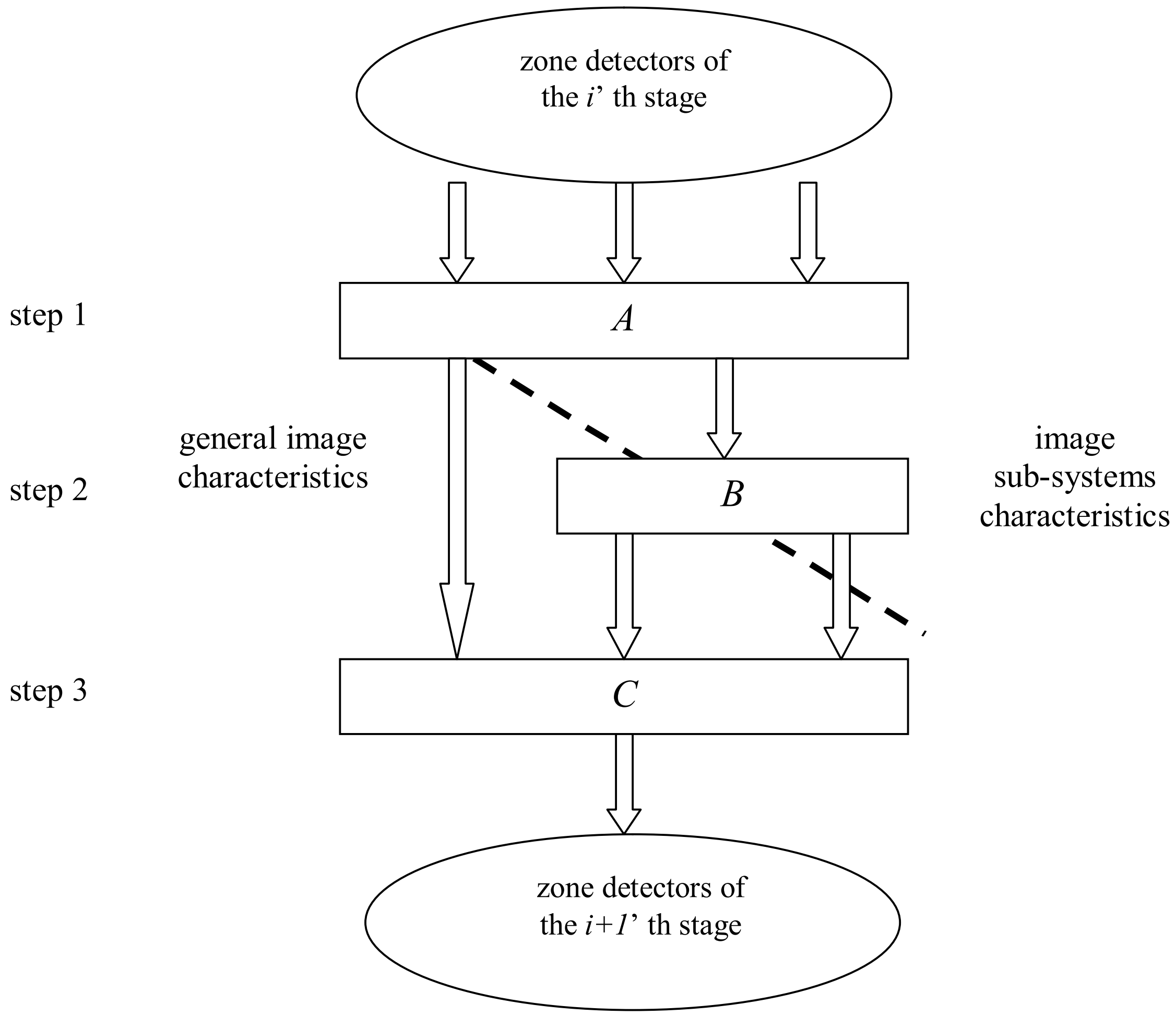

Fig. 8. Interconnection of steps at the stage of data processing

## 3. Conclusions

Currently, there are two classical approaches to the construction of intelligent systems (IS) based on artificial neural systems (ANS) and based on semantic networks (SN). IS based on ANS are practically designed only for recognition of images in a narrow domain. FIS PSS are designed not only for universal recognition of image elements (associative recognition), the movement, the whole image, scenes and events, i.e. processes extended in time, but also for internal representation (presentation) of objects of the world. These presentations are needed to verify the hypotheses generated by the FIS RS. If IS based on SN models the process of reasoning in a natural language or a language close to the natural, then FIS RS are designed to generate hypotheses, FIS PSS learning with a "teacher" on the basis of communicative functions.

This paper refutes the idea of FIS building only on the basis of RS, SN can serve as a simplified model of this system, or only on the basis of PSS construction, as a simplified model which could be considered as ANS.

For the first time modal and vector theory describes a unified approach to the construction and operation of FIS PSS and FIS RS.

Although this paper considers FIS PSS only for presenting 2D objects of CE, the results of the research are valid also for presenting 3D objects of the world. To do this, it is sufficient to enter spatial characteristics of perception depth that defines the distance between FIS PS and an object or an element of perceived object (in the systems with binocular vision, this characteristic is defined by the mechanism of disparity), as well as to define the co-direction of collinear vectors of the change of values of spatial characteristics of perception of recognized patterns substructures. The inner surface texture of 3D object, different levels of illumination, and surface boundaries within the outer contour of the object can determine the direction of these vectors. In the considered FIS presentation of 2D objects with internal structure is performed on the same level of information processing as the presentation of scenes.

A diversity of neural elements, such as zone detectors, zone identifiers, processors as well as a complex structure of their interconnection at every stage of data processing including both ascending and descending (control) data flows described in this paper construct the model of brain neuron structure that makes it possible to explain different types of neurons, layers, the structure of their interconnection, and its flexibility that exist in the cortex of cerebrum.

**References**

[1] Gerald M. Edelman, Vernon B. Mauntcastle. The mindful brain. Cortical Organization and the Group-Selective Theory of Higher Brian Function. // The MIT Press, Cambridge, Massachusetts, and London, England, 1978, 133 p.
[2] David H. Hubel. Eye, brain and vision. // Scientific American Library a Division of HPHLP New York, 1988, 256 p.
[3] Raymond M. Smullyan. Theory of formal system // Revised edition Princeton New Jersey, Princeton University Press, 1962, 207 p.
[4] Stephen Cole Kleene. Mathematical Logic // John Wiley & Sons, Inc. New York, London, Sydney, 1967, 480 p.
[5] Fernandez-Duque, D., Johnson, M. L. Attention metaphors: How metaphors guide the cognitive psychology of attention. // Cognitive Science, 1999, Vol. 23 (1), p. 83 – 116.
[6] Yu.V. Parzhin. Fundamentals of the theory of intellectual type formal systems. Structural neural network. // Systems of Information Processing. – Kharkiv: KhUAF, 2010, Vol. 6 (87), p. 2 – 12., http://www.nbuv.gov.ua/portal/natural/SOI/2010_6/Index.htm